\documentclass[letterpaper, 10 pt, conference]{ieeeconf}  % Comment this line out
\IEEEoverridecommandlockouts                              % This command is only
\usepackage{graphicx}
\usepackage{subcaption}
\usepackage{amsmath}
\usepackage{graphicx}
\usepackage{caption}
\usepackage{subcaption}
\usepackage{placeins}
\usepackage{float}
\usepackage{balance}
\usepackage{lipsum}

\title{\LARGE \bf
Design and Development of Underwater Vehicle: ANAHITA
\\
\textnormal{Conceptual Design Report: Team AUV-IITK}
}

\author{Akash Jain, Manish Kumar, Rithvik Patibandla,\\ Balamurugan R, Naveen Chandra R, Abhinav Arora, Akash K Singh,  Varun Pawar,\\ Aditya Rai, Medha Agarwal,  Priank Prasad, Vandit Sanadhya, Prateek Yadav, \\ Inshu Namdev, Nilay Shah, Saksham Mittal,  Ayush Gupta and Naman Agarwal\\ Faculty Advisor: Dr. Mangal Kothari, Department Of Aerospace Engineering, IIT Kanpur}

\begin{document}
\maketitle
\thispagestyle{empty}
\pagestyle{empty}

\begin{abstract}

Anahita is an autonomous underwater vehicle which is currently being developed by interdisciplinary team of students at Indian Institute of Technology(IIT) Kanpur with aim to provide a platform for research in AUV to undergraduate students. This is the second vehicle which is being designed by AUV-IITK team to participate in 6\textsuperscript{th} NIOT-SAVe competition organized by the National Institute of Ocean Technology, Chennai. The Vehicle has been completely redesigned with the major improvements in modularity and ease of access of all the components, keeping the design very compact and efficient. New advancements in the vehicle include, power distribution system and monitoring system. Anahita's sensors include the inertial measurement units (IMU), hydrophone array, a depth sensor, and two RGB cameras. The current vehicle features hot swappable battery pods giving a huge advantage over the previous vehicle, for longer runtime. 

\end{abstract}

\balance

\section{INTRODUCTION}
Autonomous underwater vehicles (AUVs) have the potential to revolutionize the way we explore the sea. AUVs are involved in a number of maritime areas such as maritime security, oceanography, submerged structure inspection and maintenance, climate change assessment and marine habitat monitoring. The main aim of this project is to develop a robust vehicle capable of mimicking real-life mission scenarios in an underwater vehicle.\par
The main focus of this report is placed onto robust mechanical design of an AUV with a smart power distribution system, and designing better controls and vision algorithms which can improve the stability in its manoeuvres.\par
The later sections describe the background of the vehicle which is being developed, followed by an analysis of vehicle's mechanical design and stability. The PCB designing, integration of different sensors and the controller design, along with mission planner and vision algorithm are also discussed.

\section{DESIGN OVERVIEW}
Anahita is an improvement over AUV-IITK's previous vehicle Varun in terms of its modularity, robustness, ease of manufacturing and assembly. Rigorous Finite Element Analysis(FEA) is done to improve the robustness and use of Computer Numeric Control(CNC) machining is increased. Some components are interchangeable and work specific parts increases the vehicle's modularity significantly. The design is similar to that of the Remotely operated vehicle (ROV). The vehicle is designed to perform complex space-constrained tasks and at the same time, not compromising on the maneuverability.

Major achievements in the electrical subsystem are custom design and fabrication of on board electrical circuit boards. A significant improvement in the electronics has been made by implementing modular design, which helps in debugging the system a lot faster. A full suit of inertial, visual and acoustic sensors are being used for data collection and navigation. The vehicle is powered by two lithium-ion polymer batteries, which have higher specific energy.

\begin{figure}[h!]
\includegraphics[width=\linewidth]{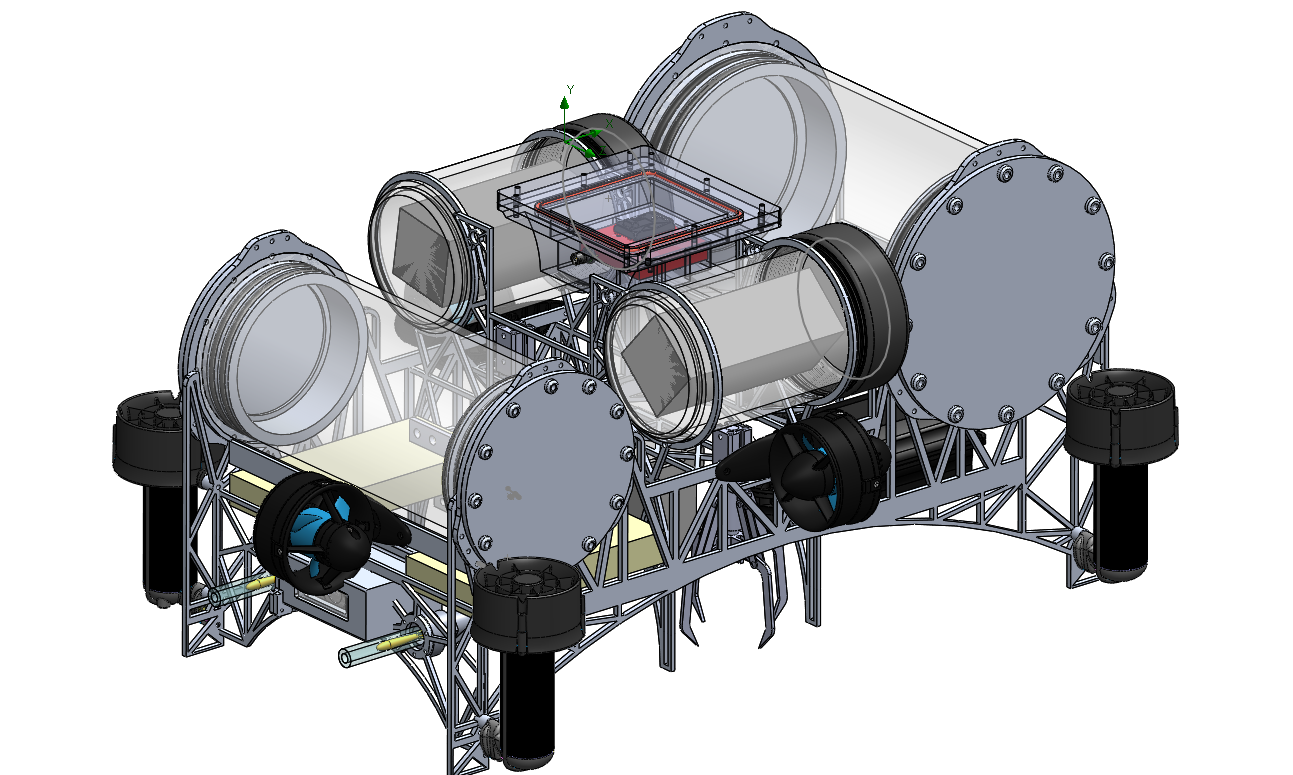}
\caption{Anahita Vehicle}
\end{figure}

Another important challenge was to develop low level controllers which will be capable of producing stable motion along all the degrees of freedom that our vehicle provides, so that we can have strong base upon which we can build our motion layer for the vehicle. Next step was to focus on individual tasks that are to be performed in the competition. We planned to write and test the code side by side so that we trace back the error, if any. After that the tested code had to be formatted in a proper way, using all the functionalities provided by ROS like action client, servers, publishers, subscribers, etc. Development of vision layer was decoupled from motion layer. We saved all the underwater camera data and used it to test the different vision algorithms. The aim of the vision layer is to provide motion layer with stable and accurate information about the objects present in the arena. So, it first enhances the degraded underwater image and undergoes some noise reduction techniques, taking care of the amount of computation going on behind it.

\section{MECHANICAL} 

The mechanical subsystem if responsible for design and manufacture of the vehicle. Anahita's mechanical system consists of the vehicle's frame, grabber, marker dropper, torpedo, connectors and penetrators.
 \subsection{Frame}
 
The frame has been completely redesigned to provide a strong structural support and also to hold all the components together in place. One of the main challenges in the frame design, was to reduce the mass, by keeping the structural rigidity same. Unlike our previous vehicle VARUN, the current frame is a wired frame with truss structures, to provide structural support and also reducing the mass.
The basic design of the frame was made and ANSYS Topology optimization was run in order to reduce the mass. The solver output was set to give 20\% of mass of the original input design.\newline
\begin{figure}[h!]
\includegraphics[width=\linewidth]{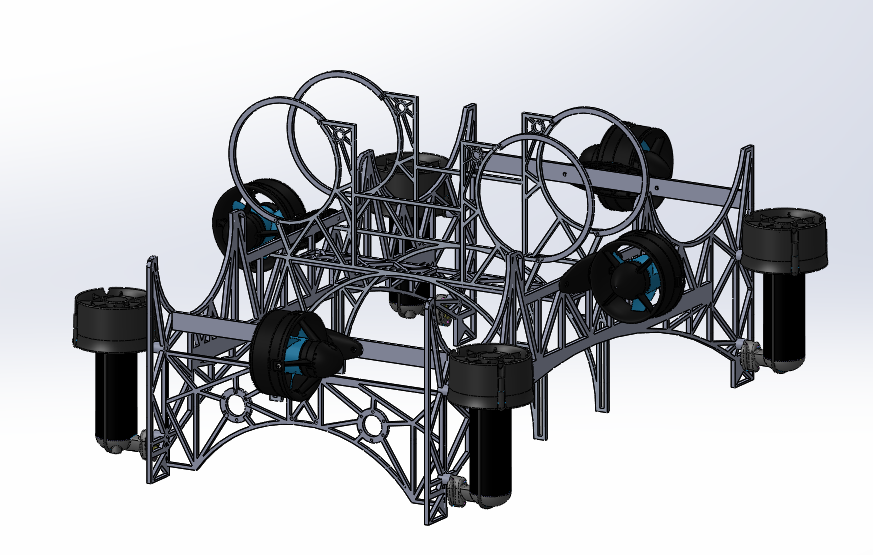}
\caption{Frame of Anahita}
\end{figure}
The frame is made by Abrasive Water-jet Machining, using Aluminium 6061-T6 Alloy. Aluminium sheets of two different thickness are used. 5mm thick sheets are used for the side plates and 4mm thick sheet for the rest of the frame.
The frame also supports the eight thruster configuration, which the vehicle is currently using.

\subsubsection*{Advantages}
All the components are attached directly to the frame, so separate mounts are not required for individual parts. The above design reduces weight, complexity and allows for efficient use of space. All components are placed such that the center of mass is at the relative center. This provides directional stability and ease in control.

\subsection{Watertight enclosures}
These housings provide watertight protection for the components like electrical circuits, actuator controls, batteries and sensors. Nitrile O-rings are used for sealing. All the grooves for the O-rings have been made using the MACRO Rubber and Plastic Guide. The main objective for our current vehicle was to make modular design for all the enclosures, keeping in mind the ease of mounting mechanism for the enclosures.

\subsubsection{Camera Casing}
The Camera casings provide watertight protection for the two cameras, which are employed for the front view and the bottom view individually. The camera being used in the vehicle has a 90$^{\circ}$ field of view. It has been ensured in the design that the view of the camera is not obstructed. An acrylic opening has been made in the middle, to allow for the camera view. The entire casing has been manufactured using Aluminium 6061-T6 Alloy, by  Computer Numeric Control (CNC). A face O-ring seal is used for providing waterproof support. The O-ring is pressed by a 5mm thick Aluminium plate, fastened with Stainless Steel Allen bolts in order to provide adequate force for sealing.\newline
\begin{figure}[h!]
\includegraphics[width=\linewidth]{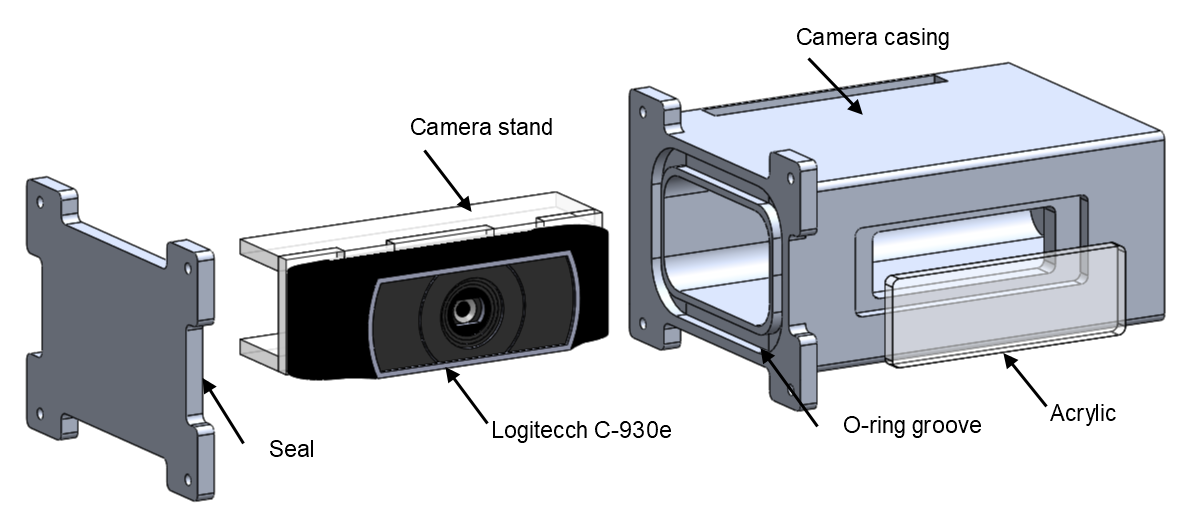}
\caption{Exploded view of camera casing}
\end{figure}
 The camera is being supported by acrylic plates, for maintaining orientation inside the camera. The entire design has been made very compact, leaving no room for wasted space compared to the design in the previous vehicle VARUN. Fischer Connectors are being used as a means for the incoming signal wire to enter the casing which are placed on the Aluminium lid used for sealing.

\subsubsection{Main Hull and SOV casing}
The main Hull houses all the electronics of the vehicle. The main Hull along with the SOV casing are the main source of buoyancy. The Hull is made of three parts- Acrylic tube, Flange and the Cap. The Cap and the Flange are made from Aluminium 6061-T6 Alloy, manufactured using CNC lathe and milling methods. The Acrylic tube acts as a transparent interface for the monitoring of all the electronic displays and the LED indicators. 

The sealing method of the main Hull and the SOV casing has been significantly improved over the previous vehicle, discarding the use of epoxy to fill the gap between the Acrylic and the Flange. This improvement is made by using two static radial O-rings for preventing water leakage. There is also a face O-ring seal between the Flange and the Cap. The necessary compressive force for the O-rings are provided by using M5 threaded holes on both Cap and Flange. The electrical connections between the Hull and the batteries, actuator and sensors outside are made using Fischer connectors. These are very rugged connectors, which has a depth rating of around 100m. 

The SOV casing is designed similar to the Hull, but a little smaller in size. It houses an air cylinder, pressure regulator, two 3/2 valves and two 5/2 valve, for the pneumatic actuators present in the Grabber and the Torpedo. The air connections between the casing and the actuators outside has been made completely modular, by using Bulkhead Air Connectors. 

\begin{figure}[h!]
\includegraphics[width=\linewidth]{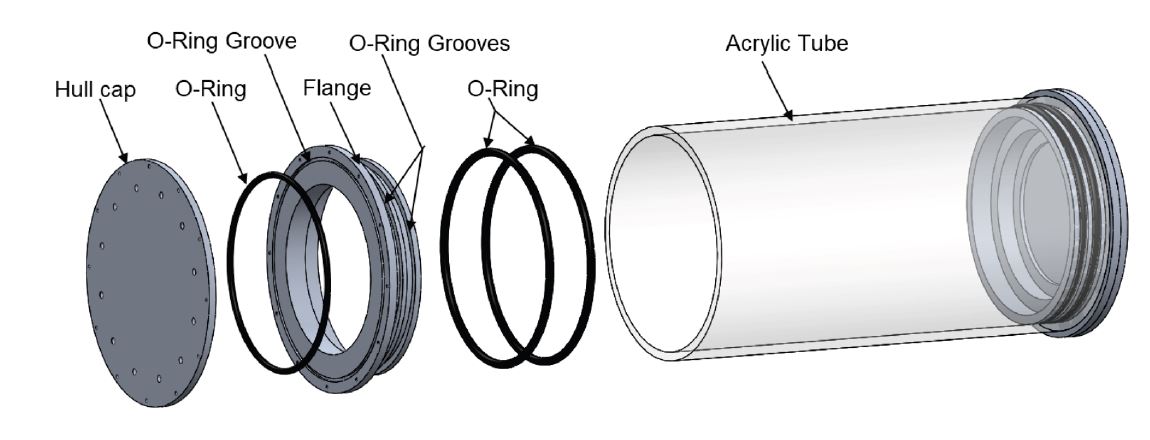}
\caption{Exploded view of Main Hull}
\end{figure}

\subsubsection{Battery Pod}
In our previous vehicle, all the batteries were placed inside the hull which made it complicated for access and hard to charge. This process cost us a lot of time. Therefore, we are moving to an external battery pod system using two battery pods, each containing one battery, that will free up space in the new hull and make charging and accessing the batteries easier and safer.

The design of battery Pod consist of two parts- Outer body of Battery Pods and Battery pod Internals. The outer body of battery pod consist of an Acrylic tube which is closed at one end. The open end of pod is threaded to fit the  battery pod lid for quick battery removal and replacement. The new battery pod is 20 cm long, with an OD of 130mm and ID of 114mm each with a thickness of 8mm which is sufficiently robust and large enough to hold the batteries. Each battery pod features two Fischer connectors, one 2 pin connector for  Power supply, and an 8 pin connector for charging and discharging the batteries. These circular connectors are fixed on the closed end of acrylic hull. Each battery pod houses a 22.2V 10000mAh Tattu battery.
\begin{figure}[h!]
\includegraphics[width=\linewidth]{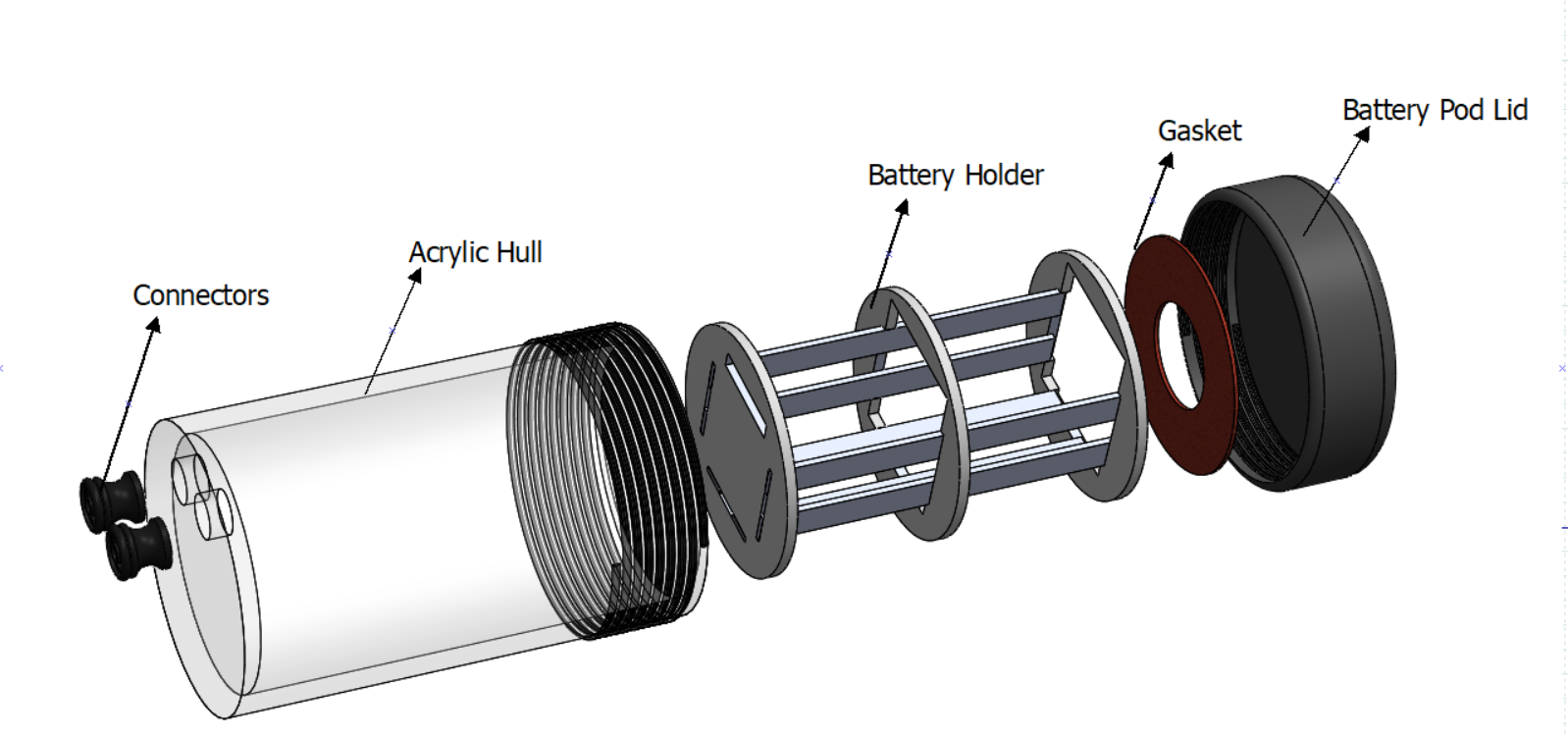}
\caption{Exploded view of Battery Pod}
\end{figure}

The internals consist of a battery holder consisting of three nylon plates mounted inside the slits of three nylon circular disc that fits exactly inside the circular hull. 2mm gaskets are being used for watertight sealing of the battery pods. The gasket is placed as a layer between the lid and the acrylic tube. The necessary force for the compression is achieved by the rotation of the lid over the threading present on the outer face of the acrylic tube.
\subsubsection*{Operational Advantages}
The new Battery pods are Hot Swappable i.e we can swap individual battery pods without affecting the operation of the system. The whole body of Battery pods is made of Acrylic (PMMA) that is a transparent light weighted thermoplastic having high tensile strength, flexural strength, transparency, chemical resistance and heat resistance. Batteries can be charged and discharged directly without any removal from pods. This eases recharging process and makes the design highly modular.

\subsubsection{IMU casing}
The inertial measurement unit uses gyroscope, accelerometer and magnetometer to localize the vehicle with respect to the initial position. It takes Earth`s magnetic field as reference values to measure the change in angular position during vehicle motion. Any disturbance in the magnetic field would cause the IMU to generate errors in the values given as output. As a countermeasure, the IMU is kept in a separate watertight casing away from components like power distribution board and solenoid valves. 

The casing is milled entirely out of a single Acrylic block. The sealing is ensured using Nitrile O-rings and the acrylic cap provides the necessary force for the O-rings, using fasteners. The sensor is placed at a certain height in order to prevent any damage from accidental water-leakage. Separate mounting points are provided for both the developer board and the IMU, for easy assembly and disassembly. Fischer connectors are used in order to make the casing completely modular. 
\begin{figure}[h!]
\centering
\begin{subfigure}[b]{0.4\linewidth}
  \includegraphics[width=\linewidth]{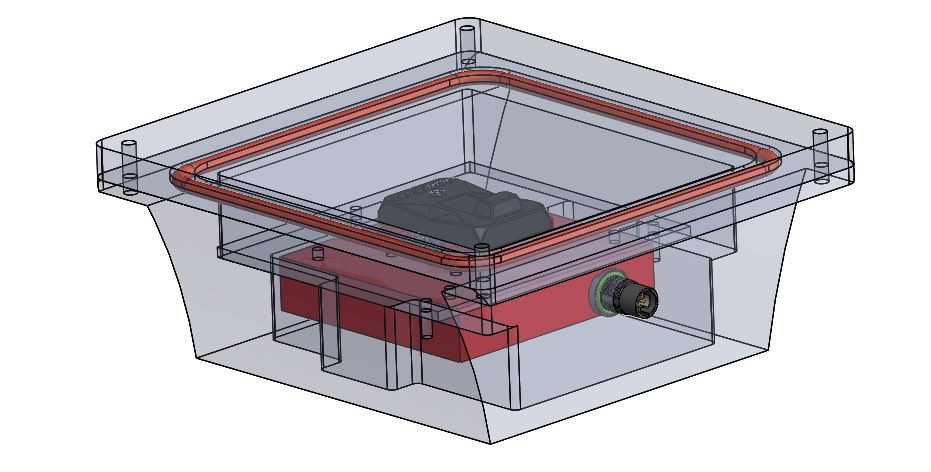}
  \caption{IMU Casing}
\end{subfigure}
\begin{subfigure}[b]{0.35\linewidth}
  \includegraphics[width=\linewidth]{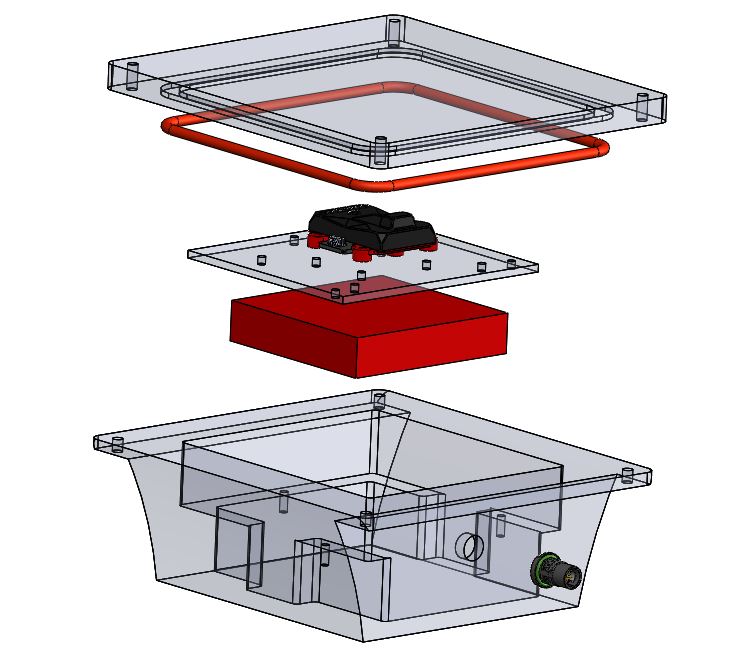}
  \caption{Exploded view}
\end{subfigure}
\caption{The watertight enclosure for Sparton AHRS-8}
\end{figure}

\subsection{Grabber}
\subsubsection*{Components and Mechanism}
The lifting action is performed by a scissor mechanism which can provide a total maximum elongation of 300mm. One end of the scissor is actuated linearly using the 5 cm piston and the other end is kept fixed. A mount attached to the lower end of the scissor serves as a point of attachment of the main grabber fingers. Aligned vertically and attached to the lower mount is a plate which ensures that the grabber moves only vertically and does not perform a swinging action about its point of mounting. 

The grabbing mechanism also employs four finger mechanism. The entire mechanism is actuated pneumatically by two pistons. The piston providing actuation for lifting mechanism situated on the top of scissors gives an elongation of 5 cm whereas the lower one responsible for grabbing action of the fingers, elongates by 1 cm. Both the pistons are connected to the gas cylinder using regulator and 5/2 Solenoid Valves. An object is picked and held by the grabber by manipulating the fingers.
 
\begin{figure}[h!]
\centering
\begin{subfigure}[b]{0.4\linewidth}
  \includegraphics[width=\linewidth]{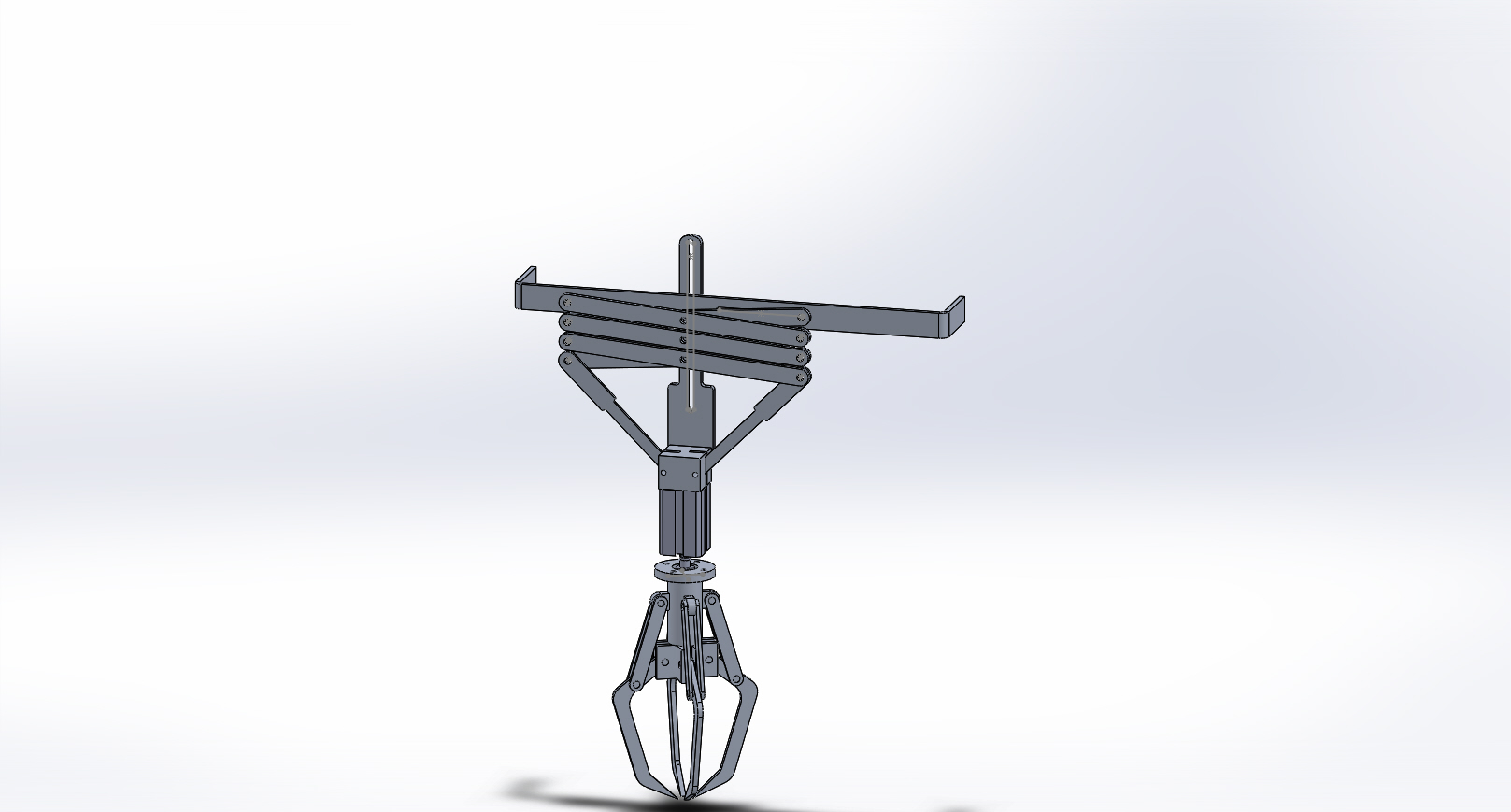}
  \caption{minimum elongation}
\end{subfigure}
\begin{subfigure}[b]{0.4\linewidth}
  \includegraphics[width=\linewidth]{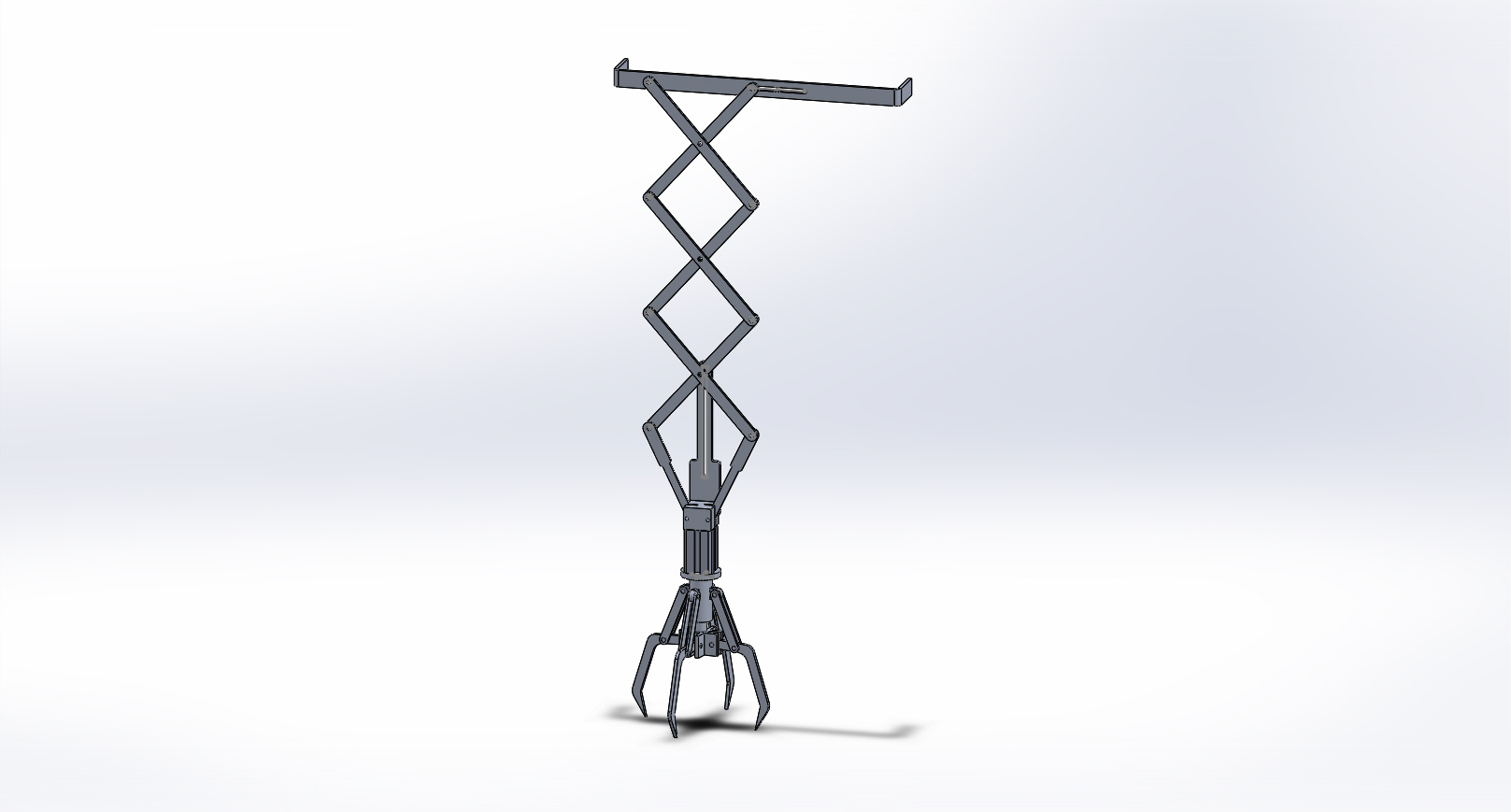}
  \caption{maximum elongation}
\end{subfigure}
\caption{Scissor mechanism}
\end{figure}

\begin{figure}[h!]
\centering
\begin{subfigure}[b]{0.4\linewidth}
  \includegraphics[width=\linewidth]{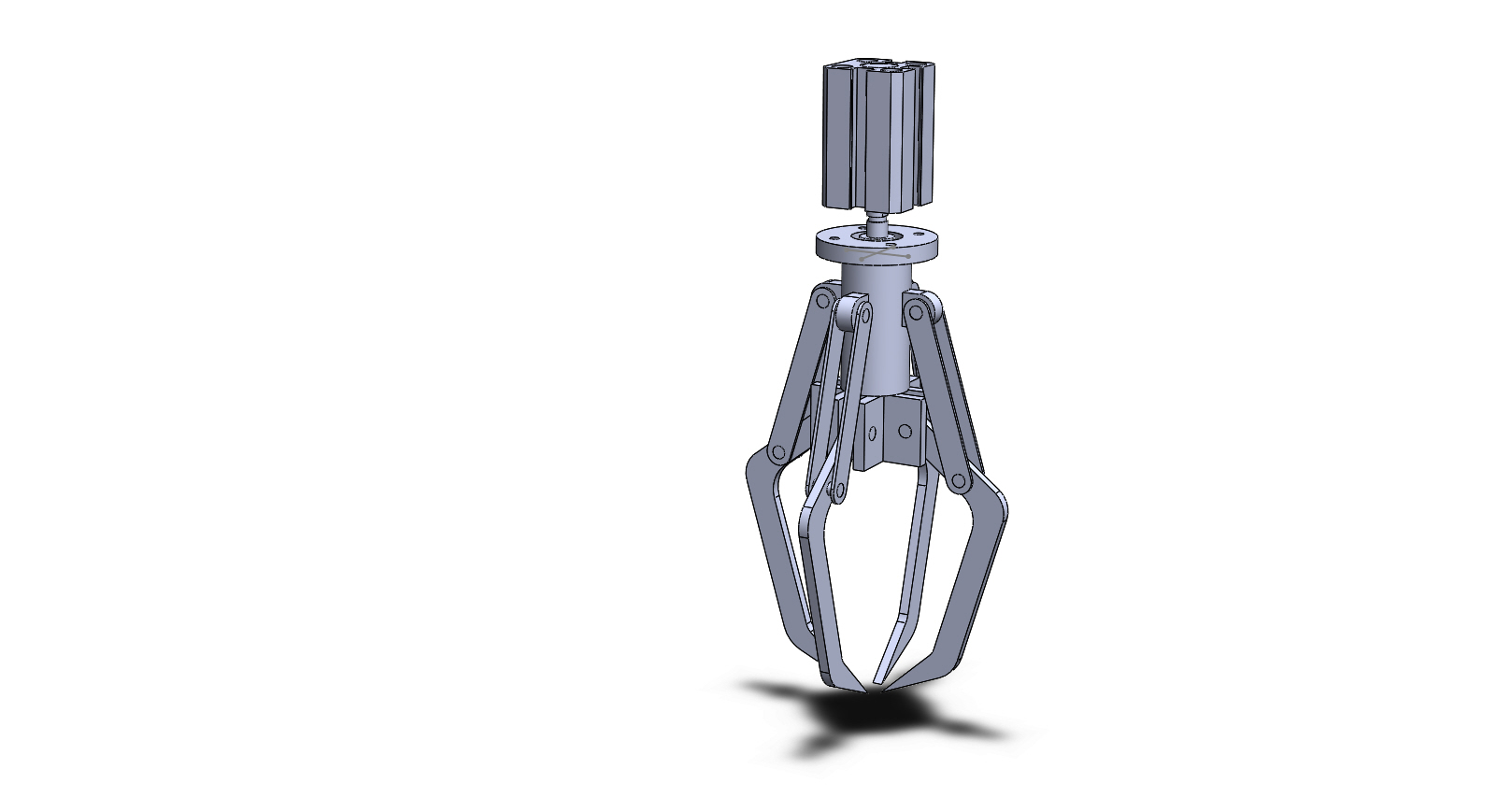}
  \caption{minimum separation}
\end{subfigure}
\begin{subfigure}[b]{0.4\linewidth}
  \includegraphics[width=\linewidth]{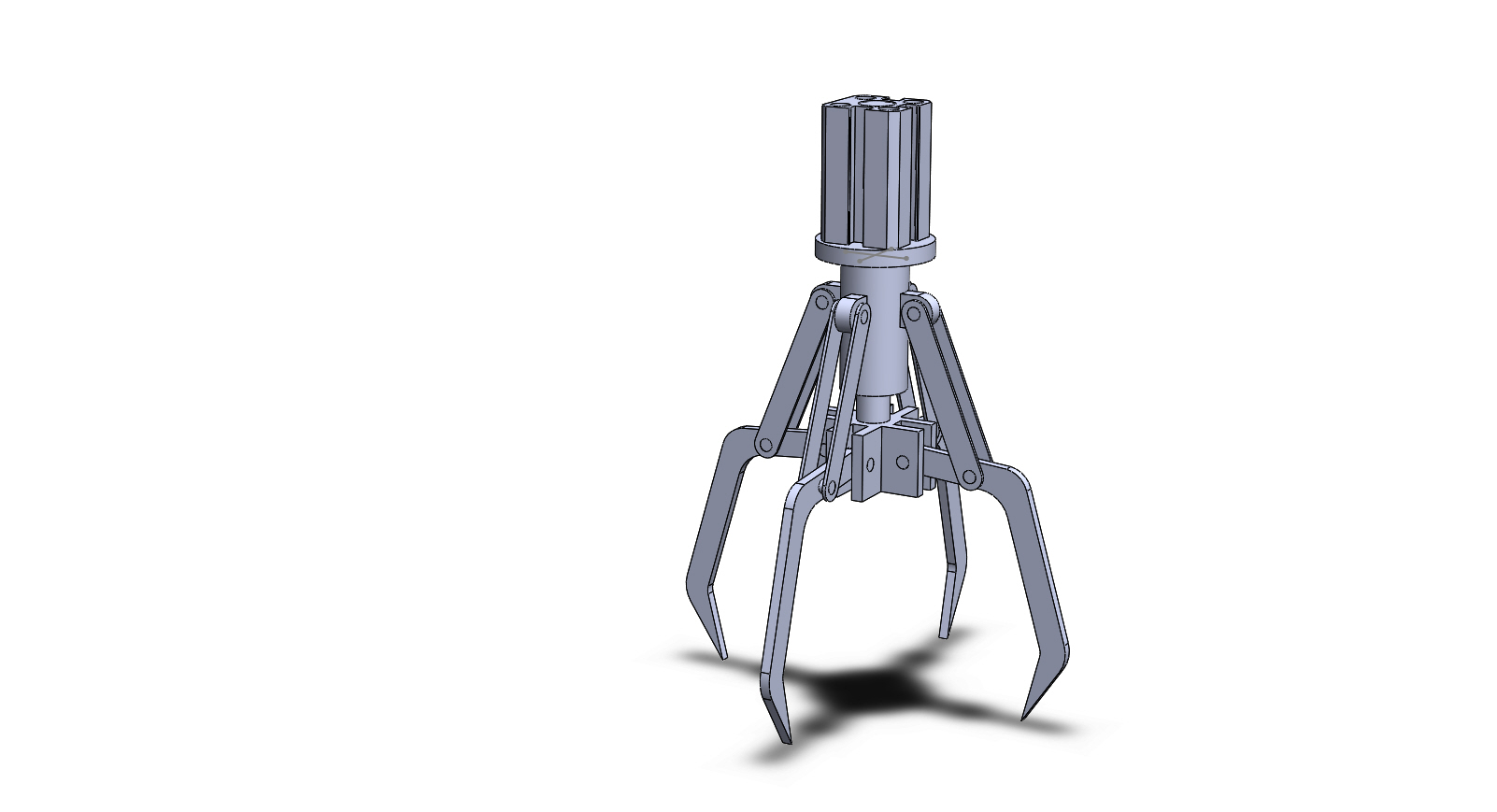}
  \caption{maximum separation}
\end{subfigure}
\caption{Grabbing action}
\end{figure}

\subsubsection*{Operational Advantages}
The tedious task of picking objects located on ocean bed, carrying them and then placing them back is highly facilitated by this part of the vehicle. The scissors mechanism is highly proficient as it provides an extension of 30cm from just a linear actuation of 5cm and the grabber achieves a maximum diagonal extension of 120mm with a 1cm actuation.

\subsection{Marker Dropper}
The markers (balls) used here are golf balls of diameter 45mm. The dropper prototype is capable of holding two balls and dropping them independently. The dropper is actuated using a waterproof servo. The prototype mainly consists of two parts, the outer cylinder which is clamped onto the frame and the inner cylinder which contains the two balls. The design exploits the non-concentric nature of the cylindrical hole made into the the two components which helps in dropping the balls at the desired locations. It can be easily placed in a cuboid of dimensions 11cm x 7cm x 7cm showing its compactness. It can be easily assembled and disassembled due to its modular design. The mechanism, being actuated using a servo instead of pneumatics, is more precise and accurate since the servo can rotate it by any angle precisely. The Dropper has been manufactured by Rapid prototyping, using PLA as the additive material.
\begin{figure}[h!]
\centering
\begin{subfigure}[b]{0.4\linewidth}
  \includegraphics[width=\linewidth]{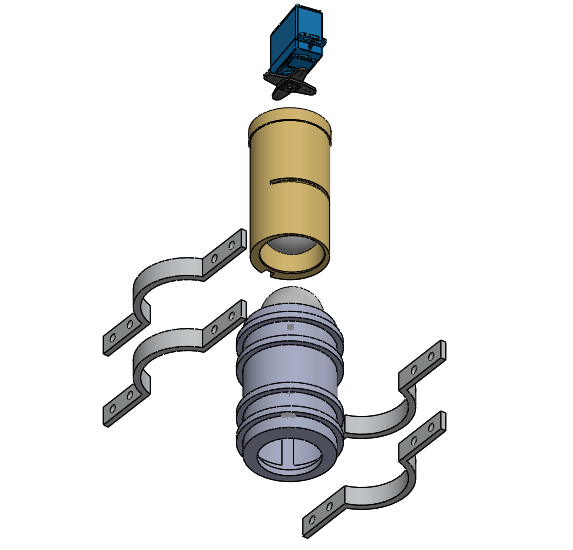}
  \caption{Exploded View}
\end{subfigure}
\begin{subfigure}[b]{0.35\linewidth}
  \includegraphics[width=\linewidth]{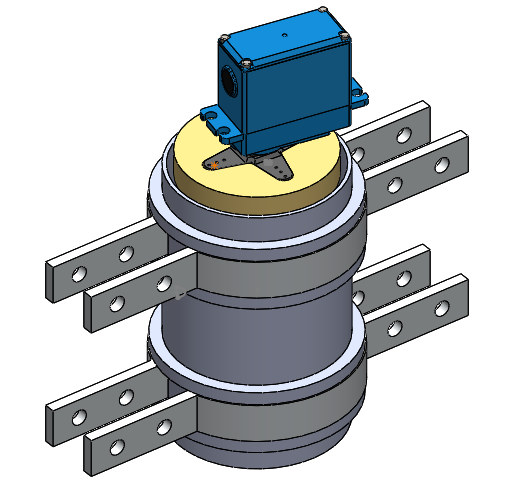}
  \caption{Isometric View}
\end{subfigure}
\caption{Marker Dropper}
\end{figure}

\subsubsection*{Working}
The outer cylinder has a non-concentric cylindrical hole of diameter 45 mm at its base which makes for the passage of the ball. On its inner wall, it also has a ledge(a protrusion) which helps in preventing the second ball from dropping while the first ball is being dropped. The inner cylinder also has a non-concentric cylindrical hole throughout its length of diameter 45 mm. It is this hollow passage which contains the two balls. Added to this, the inner cylinder has a slit along its outer circular surface in which the ledge of the outer cylinder is placed. As the servo motor rotates the inner cylinder, the hole containing the balls gradually coincides with the non-concentric hole in the outer cylinder base. The moment these two holes coincide, the first ball to drop. The ledge is positioned in such a way that it obstructs the motion of the upper ball only at the time of coincidence of the two holes. So, at this instant, the ledge stops the second ball from being dropped. The second ball is dropped in a similar fashion. This design allows for dropping the two markers at desired positions with only one actuator.

\subsection{Torpedo}
\begin{figure}[h!]
\includegraphics[width=\linewidth]{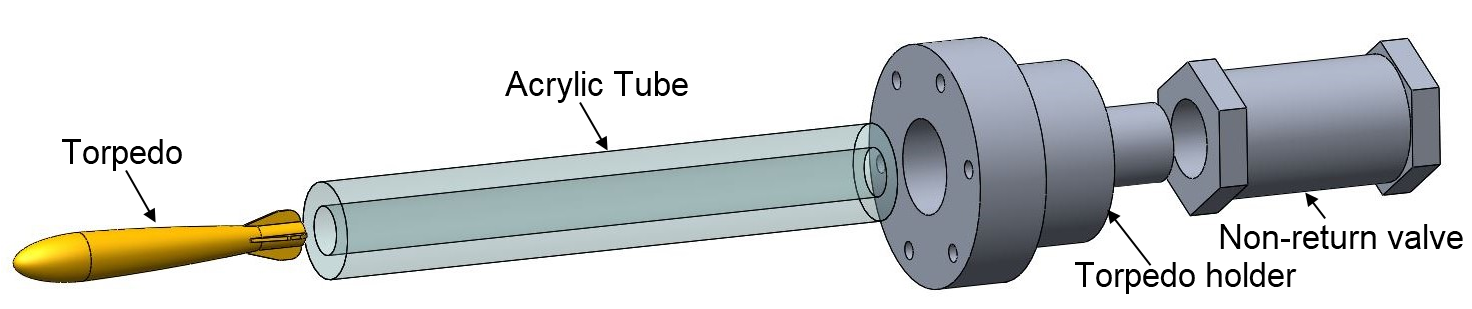}
\caption{Torpedo Launching setup}
\end{figure}

The torpedo is 3D printed using PLA. Its average density is slightly less than of water. Being positively buoyant it can be recovered easily even after it has been fired underwater. It's streamlined design reduces drag. The center of gravity and center  of buoyancy do not create torques as they are coincident with the center of mass of the  torpedo  resulting in a highly stable design. The tail fins reduces tilting and change of direction of torpedo forcing it to move linearly until it stops. When the torpedo starts going off it's straight trajectory i.e it tilts, the fins tilt as well making an angle with the flow. This flow generates torque restoring the torpedo to it's initial position.

\begin{figure}[h!]
\includegraphics[width=\linewidth]{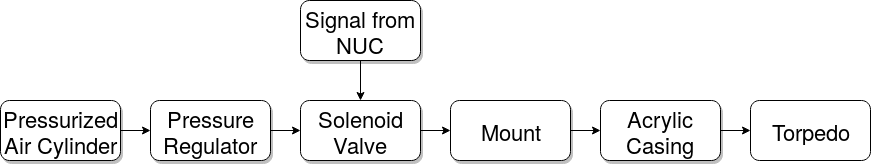}
\caption{Torpedo Launch Mechanism}
\end{figure}

 The pneumatic air cylinder contains air at 50 bar. It is brought down to desired pressure using a pressure regulator, which is then taken to SOV. When the SOV is energized using signals from NUC, It allows air to pass through it propelling the torpedo forward.
 
\subsection{Static stability analysis and Numerical model}
In Figure 12,\\ 
O: Origin\\
CG: Center of Gravity\\ 
CB: Center of Buoyancy\\
CG Coordinates: 44.92, -216.65, 21.14\\
CB coordinates: 44,92, 164.65, 21.14\\
\begin{figure}[h!]
\includegraphics[width=\linewidth]{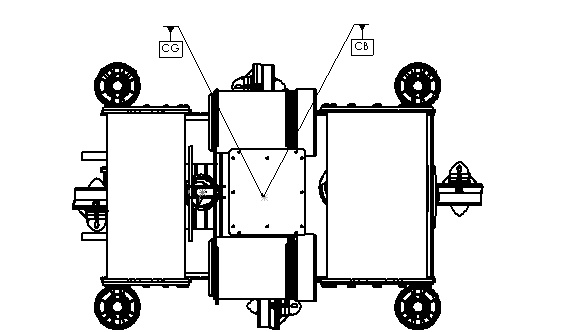}
\end{figure}
\begin{figure}[h!]
\includegraphics[width=\linewidth]{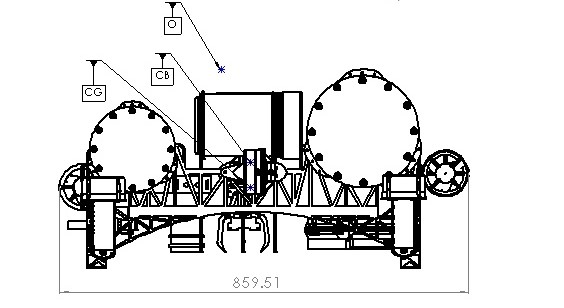}
\end{figure}
\begin{figure}[h!]
\includegraphics[width=\linewidth]{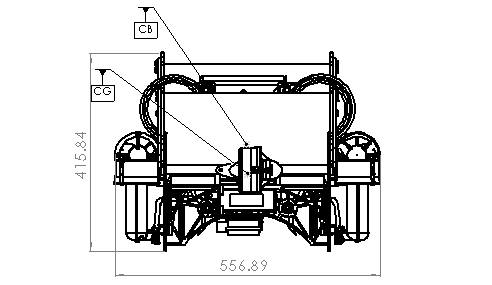}
\caption{Sketch and dimensions of Vehicle (All dimensions are in mm}
\end{figure}
Anahita consists of 8 thrusters, four for vertical direction, two for forward direction and last two for lateral direction. By using these thrusters, robot can perform 6 degree of freedom (DOF) movement (surge, sway, heave, roll, pitch and yaw). As shown in figure 13, our AUV is statically stable in yaw, pitch and roll axes as the center of buoyancy and center of gravity lie on the same line.
In the derivation of the equations that govern AUV's Numerical Model, a few assumptions were made to simplify the complex equations, they are:
\begin{itemize}
 \item AUV has constant mass and inertia tensor 
 \item AUV is designed to carry out missions in low-speed condition, hence the coupling terms can be neglected
 \item AUV is buoyant while carrying out its mission, so the buoyancy force (\(B\)) will equal to fluid mass that was displaced by robot.
 \item Missions is carried out in a shallow water environment
 \item The formula takes account of axes not coinciding with the AUV’s principal axis of inertia, %%therefore, the product of inertia is non-zero  
\end{itemize}

Newtonian method is used to model the dynamics of the underwater robot. 
$$M\dot{\nu} + C(\nu)\nu + D(\nu)\nu + g(\eta) = T$$
where,\\
\(M\) is the mass and inertia, \(C(\nu)\) is the Coriolis effect and centripetal, \(D(\nu)\) is the Hydrodynamic damping, \((\eta)\) is the gravitational and buoyancy force, \(T\) is the generalized force,

\(\nu = \begin{bmatrix} u & v & w & p & q & r \end{bmatrix}\)

\(T = \begin{bmatrix} \tau_x & \tau_y & \tau_z & \tau_\phi & \tau_\theta & \tau_\psi \end{bmatrix}\)
\newline

In order to control the robot posture, eight thrusters (\(T_n\)) (see Figure 13  for positioning of thrusters) are used to generate a generalized force of each axis (\(\tau_n\)) to control the underwater robot to convert a force of each degree of freedom to each thruster. Because the robot system is an over actuated system, thus, superposition is applied to make it easier to be calculated. Here,
\(l_i\) is the length from center of the robot to center of thruster \(i\).
\begin{center}
\(
\begin{bmatrix}
\tau_x \\ \tau_y \\ \tau_z \\ \tau_\phi \\ \tau_\theta \\ \tau_\psi
\end{bmatrix}
=
\begin{bmatrix}
1 & 1 & 0 & 0 & 0 & 0 & 0 & 0\\
0 & 0 & 1 & 1 & 0 & 0 & 0 & 0\\
0 & 0 & 0 & 0 & 1 & 1 & 1 & 1\\
0 & 0 & 0 & 0 & l_1 & l_1 & -l_1 & -l_1\\
0 & 0 & 0 & 0 & -l_2 & l_2 & -l_2 & l_2\\
-l_3 & l_3 & l_4 & -l_4 & 0 & 0 & 0 & 0\\ 
\end{bmatrix}
\begin{bmatrix}
T_1 \\ T_2 \\ T_3 \\ T_4 \\ T_5 \\ T_6 \\ T_7 \\ T_8
\end{bmatrix}
\)
\end{center}

The weight of the bot in air is 26.4 Kg and the total buoyancy is 35 kg. In order to keep our bot slightly positively buoyant we plan to add dead weights. The calculated drag force over our vehicle for surge at 0.6 m/s velocity is 10.8 N and the drag force for sway at 0.3 m/s is 6.02 N.
\begin{figure}[h!]
\includegraphics[width=\linewidth]{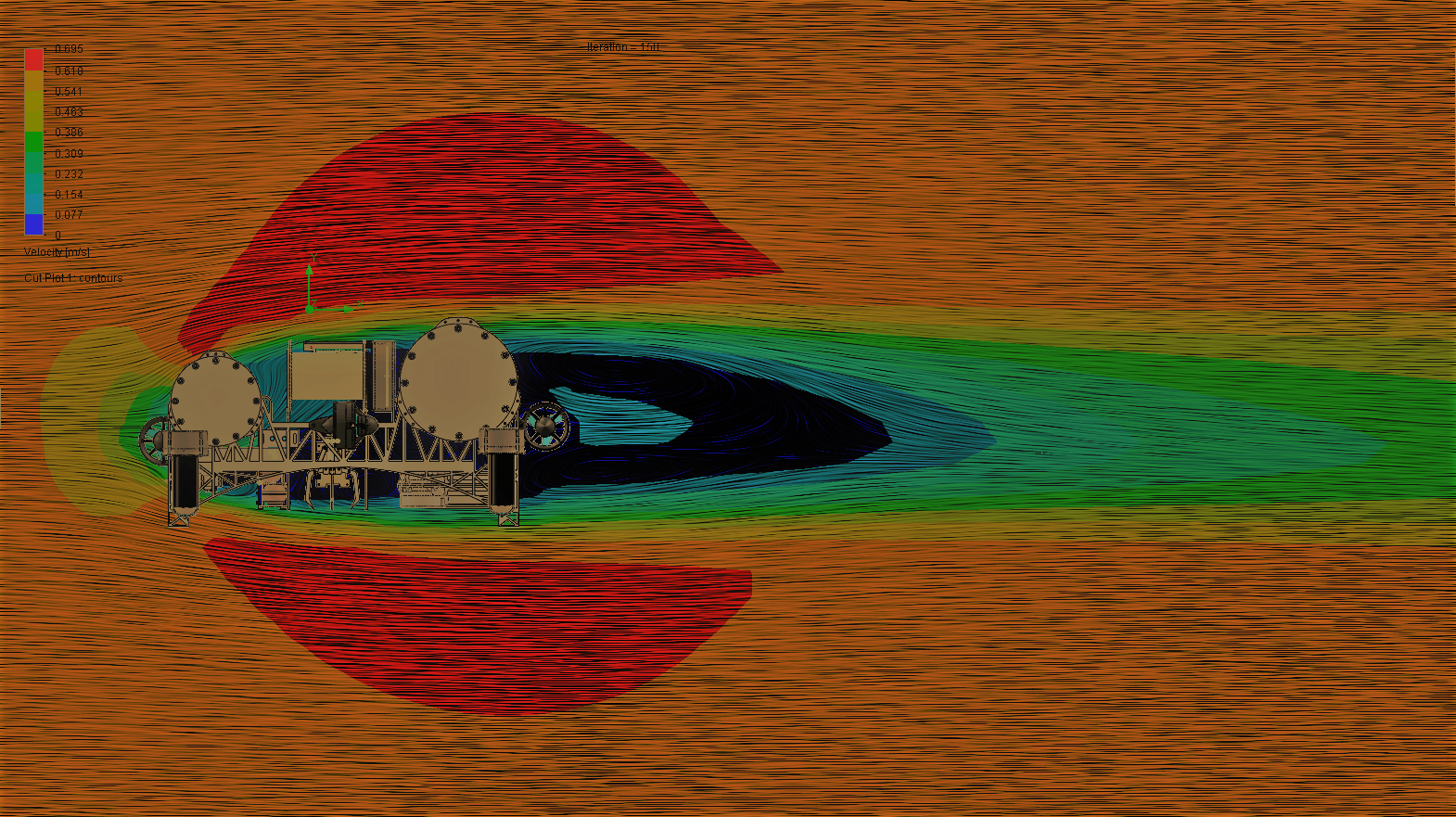}
\caption{Streamlines over vehicle for surge}
\end{figure}
\begin{figure}[h!]
\includegraphics[width=\linewidth]{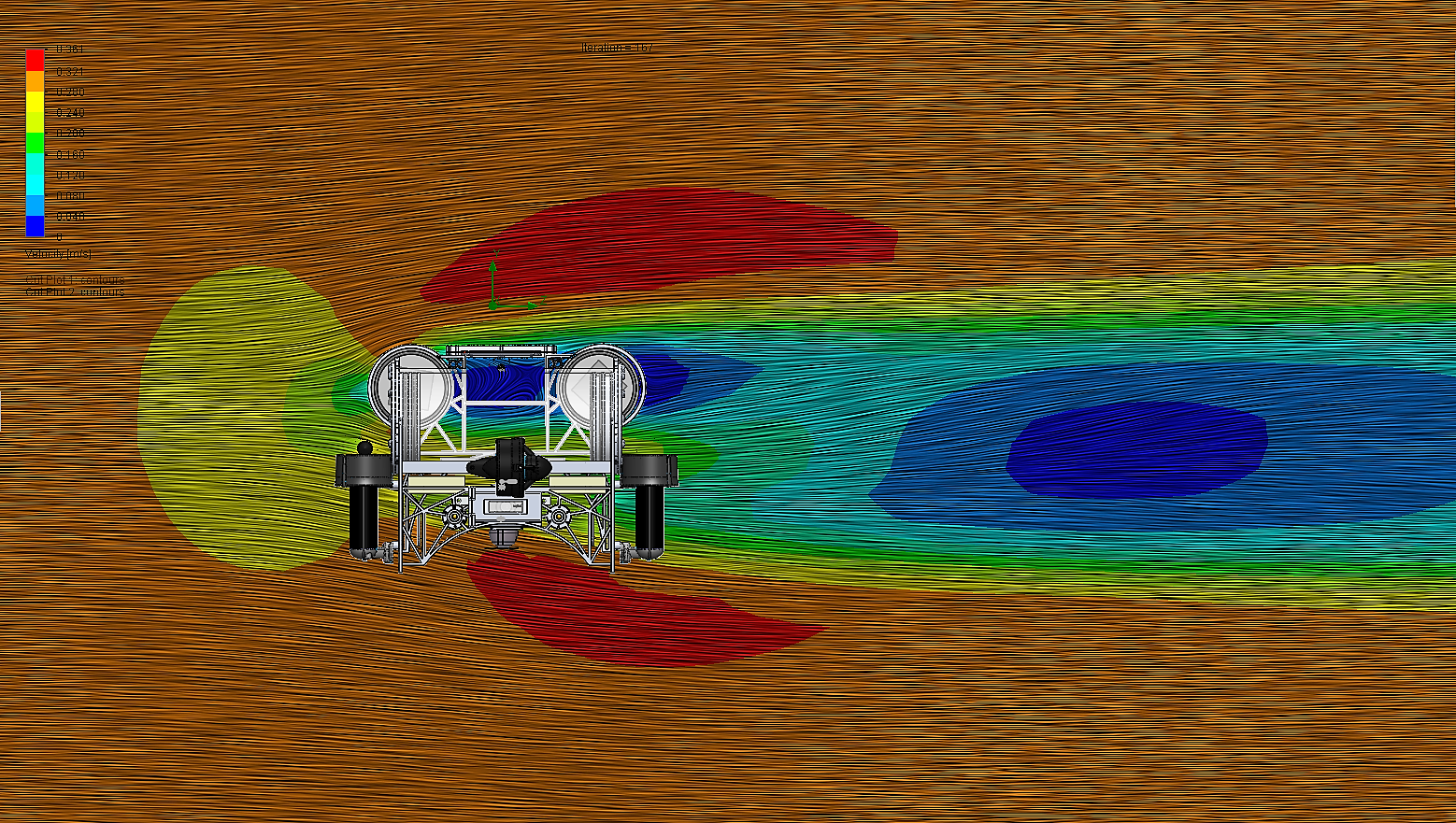}
\caption{Streamlines over vehicle for sway}
\end{figure}
  
\section{ELECTRICAL}

The electrical system in Anahita is designed for providing power, driving actuators and interfacing with various sensors installed in the robot. 
\begin{figure}[H]
\centering
\fbox{\includegraphics[width=\linewidth]{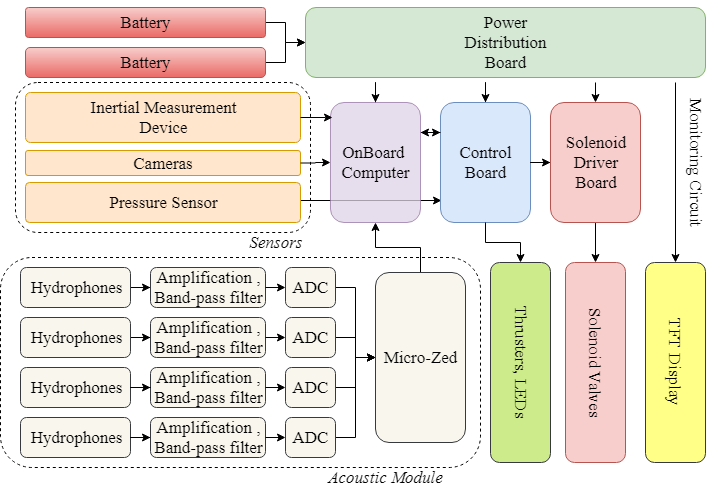}}
\caption{Complete Electrical Flowchart}
\end{figure}

The major improvement from Varun's electrical system is the custom-made PCBs which have designed to suit the specific needs of Anahita. This made the system modular and compact. There are separate PCBs for distributing power, actuating thrusters, operating solenoid valves and for processing signals from hydrophones. 

\subsection{Power}
Anahita consists of a Power Distribution System and a Monitoring System. The vehicle is powered by two 22.2V 10000mAh Lithium Polymer batteries. Li-Po batteries are chosen primarily for their high energy density but they also offer a great balance between other features such as high discharge rate, low self-discharge rate, and good cycle life.

\subsubsection{Power Distribution System}

Various components in the vehicle require input power at different voltages for their functioning. To handle this, a Power Distribution System was designed.
\begin{figure}[h]
\centering
\fbox{\includegraphics[width=0.8\linewidth, height= 5cm]{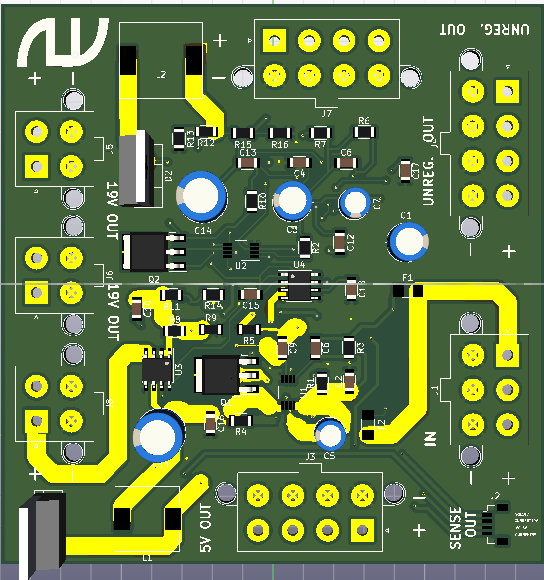}}
\caption{Power Distribution Board Prototype}
\end{figure}
The 22.2V supply was distributed into 5V, 12V, 19V regulated rails using buck converters (LM25085 from Texas Instruments). An unregulated rail was created from the batteries which powers the thrusters.

\subsubsection{Monitoring System}

The current and voltage across each rail of the Power Distribution System is monitored. The current is monitored by Hall effect current sensors (ACS725 from Allegro Microsystems).The output from the sensors is sent to the Arduino. Voltage is monitored by first using a voltage divider to convert the voltage of the rail into a value in 0-5V range and then reading the value using the built-in ADC in the Arduino.

Arduino Mega is used to control the actuation and read data from various sensors which in turn is controlled by an Intel NUC.

\subsubsection{Kill Switch}

The vehicle has two kill switches to terminate the mission in need of faults or just to use it as a switch. One switch is Hard Kill switch which completely shuts down the vehicle. The other soft kill switch stops the power to all the processes except the the power to the Onboard computer.

\subsection{Actuation}

The vehicle's motion is derived from BTD150 (from Teledyne SeaBotix) and T200 (from Blue Robotics) thrusters. The BTD150s are brushed DC motors which are operated using MC34932 Dual H-bridge ICs. The thrust is controlled by changing the duty cycle of the PWM signal sent from the arduino to the MC34932 ICs while the direction of thrust is controlled using signals sent to the direction pins of the ICs. The T200 is a brushless DC motor, which we control using Blue Robotics Basic ESC Rev 3. The thrust and its direction is controlled by varying the frequency of PWM signal sent to the ESC from the arduino.

A Solenoid Actuation board was also designed to actuate the solenoid valves used in the vehicle, using DRV102 IC (Texas Instruments). The DRV102 is being used because it helps save energy. It supplies full output for a short interval when it is triggered and then its output becomes a square wave based on the set duty cycle. This saves power as the valve only needs high voltage during the shifting of the valve while it can hold the valve in place at a much lower voltage. The valves control the servo motors of the marker dropper.

\begin{figure}[h]
\centering
\fbox{\includegraphics[width=0.8\linewidth]{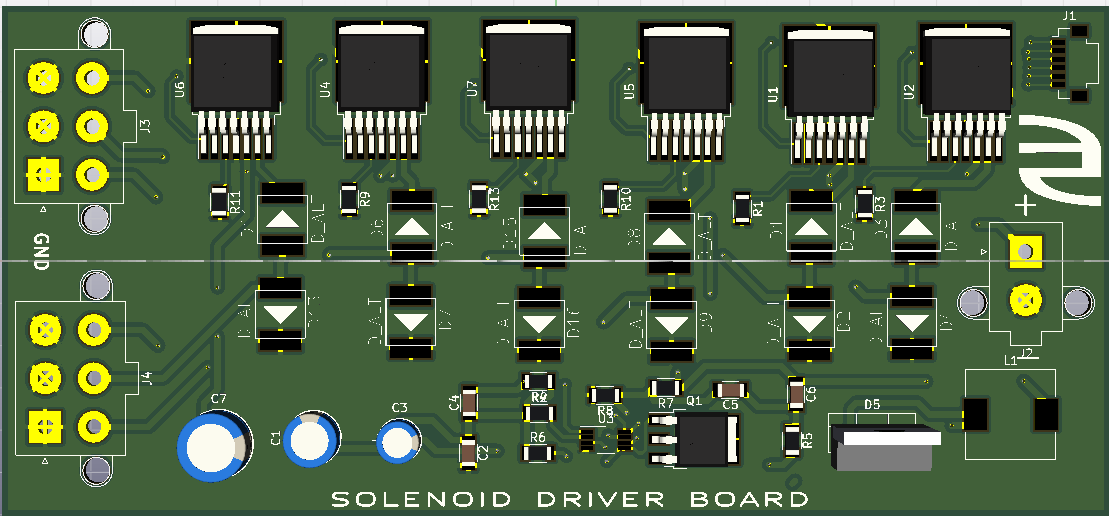}}
\caption{Solenoid Driver Board Prototype}
\end{figure}

\subsection{Hydrophones}
For the acoustic pinger signals processing, a set of 4 Aquarian Audio AS-1 hydrophones are used. Each signal received by the hydrophones is initially put through an amplifier and an analog filter. The amplifier adds a gain of fifty and sends it to the analog filter which is a 6th order low-pass filter with a cutoff frequency of 37.5 KHz. The output from the low-pass filter is re-amplified in order to map the output of filter into the input range of ADC for maximum efficiency. The signal is then converted into the differential form in order to send it to the ADC (LTC2383). The signal from the ADC is sent to an all programmable SOC development board from MicroZed which converts it into single-ended signal and then transfers it into the processor's memory. At this point, the data has been successfully transferred into the NUC and then a software level cross-correlation algorithm is applied on the processed data to find the heading of the pinger.

\subsection{Sensors}

The vehicle perceives the environment around it via various sensors such as pressure sensor and an inertial measurement unit. These signals are passed to the controller to enable the appropriate behavior. Sensors in robots are based on the functions of the human sensory organs and they require extensive information about their environment in order to function effectively.

\subsubsection{Pressure Sensor}
 The vehicle uses a bar30 pressure sensor(from Blue Robotics) which measures the depth with a resolution of 2mm and communicates via I2C protocol with the actuation board. 

\begin{figure}[h]
\centering
\fbox{\includegraphics[width=0.8\linewidth]{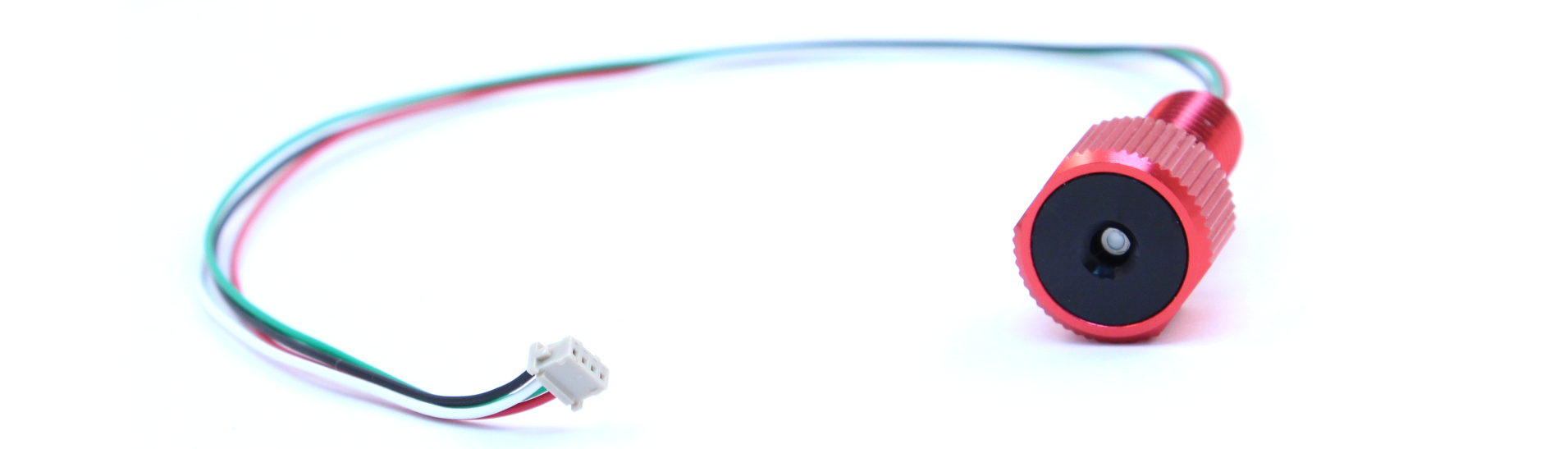}}
\caption{Logitech C930E}
\end{figure}

\subsubsection{Camera}
Vehicle needs a visual sensor which can detect and recognize object around its surroundings and take appropriate actions. ANAHITA uses two Logitech C930E cameras. These cameras come with 90$^\circ$ field of view, support 4x zoom in 1080p with auto focus feature and capable of working in low light conditions too. 

\begin{figure}[h]
\centering
\fbox{\includegraphics[width=0.8\linewidth]{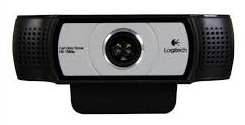}}
\caption{Logitech C930E}
\end{figure}

\subsubsection{Inertial Measurement Unit}
ANAHITA uses the AHRS-8 from Sparton as its primary Inertial Measurement Unit. It is fully temperature compensated over the operating range of -40$^\circ$ C to 70$^\circ$ C. It has a low power consumption and power management (Sleep Mode) functionality, supports multiple communication protocols and has centripetal acceleration correction.
 
\begin{figure}[H]
\centering
\fbox{\includegraphics[width=0.8\linewidth]{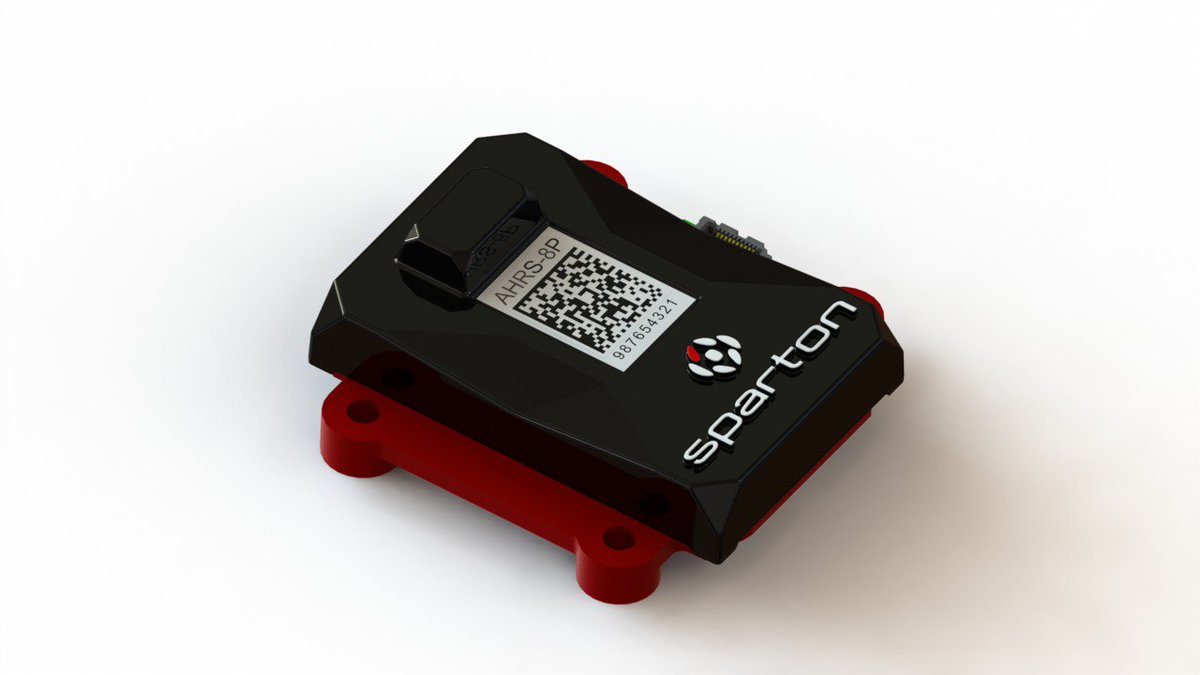}}
\caption{Sparton AHRS-8}
\end{figure}

\subsection{Connectors and Penetrators}
The electrical connections going in and out of all the watertight enclosures are made using Fischer connectors and Blue Robotics Penetrators. Connectors are mainly used for delivering the power across the enclosures. 2-pin and 8-pin connectors are used for this purpose. Using a connector for the signal wires carrying the data from sensors may result in data loss. In order to prevent this, penetrators are used to directly carry the signal wire from the sensors to the main hull, leaving no room for errors. Both the connectors and penetrators will be sealed using RS 851-044 epoxy, which is best suited for the purpose. 

\begin{figure}[h!]
\centering
\begin{subfigure}[b]{0.4\linewidth}
  \includegraphics[width=\linewidth]{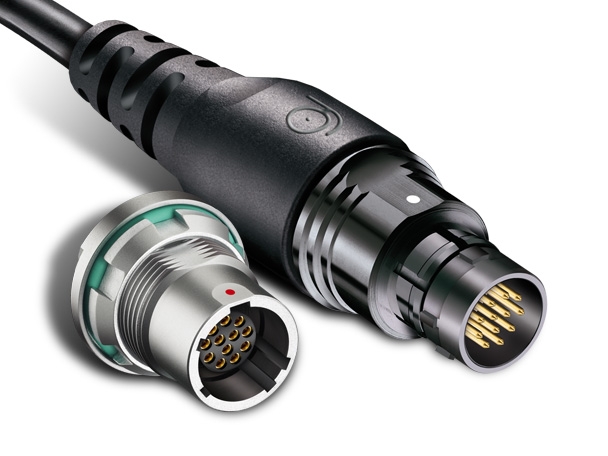}
  \caption{Fischer Connectors}
\end{subfigure}
\begin{subfigure}[b]{0.4\linewidth}
  \includegraphics[width=\linewidth]{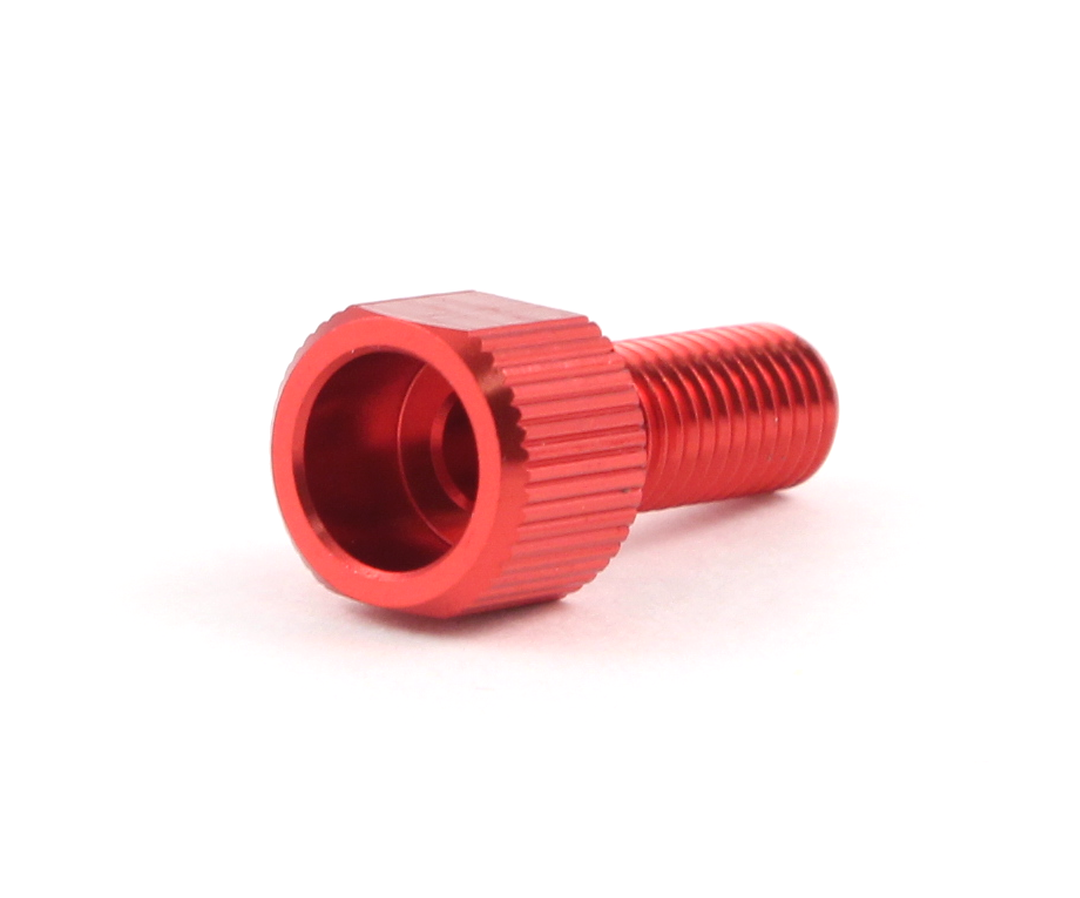}
  \caption{Penetrator}
\end{subfigure}
\caption{The Penetrators and the connectors that are being used in the vehicle for the electrical connections between the waterproof enclosures}
\end{figure}

\section{SOFTWARE}
The software stack of ‘Anahita’ consists of dedicated layers for the hardware integration, controls and navigation, motion planning, perception and acoustic localization. It is built over the Robot Operating System (ROS) framework by Willow Garage which acts as a  communication middleware between all processes running on our robot.

\begin{figure}[h!]
\includegraphics[width=\linewidth, height = 6cm]{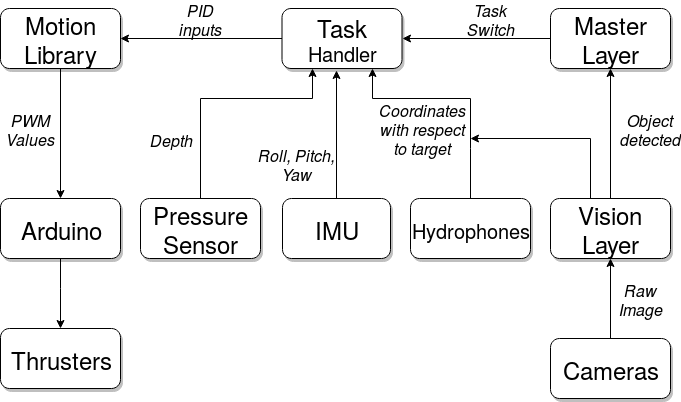}
\caption{Software Data and Control Flow}
\end{figure}

In order to make the code modular, it is divided it into five layers: 

\subsubsection{Master Layer} Master Layer is responsible to initiating each process. It instructs the task handler layer to execute nodes sequence in which the task has to be performed.
\subsubsection{Task Handler Layer} This layer has task specific code written for each task and uses motion library to achieve the target.
\subsubsection{Motion Library} This layer basically serves the task handler layer to achieve a goal. It consists of six PID control loops for six degree of freedom.
\subsubsection{Vision Layer} Vision layer is responsible for providing the information about the mission elements present in arena. It detects the targets from the raw image of the camera and provides the coordinates of the vehicle to the task handler layer.
\subsubsection{Hardware Layer} This layer is responsible for integrating sensors with the software stack. It collects all the information from the sensors and advertises it over a topic from where any node which needs it might be wanting to take it. 

The design specifications are such that it is independent and extendable. The software can scale with respect to the tasks or missions that can be accomplished.

\subsection{Controls}
Anahita has eight propulsion-based thrusters which enables it to perform motion in all the six degrees of freedom (surge, heave, sway, roll, pitch and yaw). For each of these degrees, there is implementation of an independent PID controller in our motion library.  Due to various non-idealities present between the thrusters, it was required to perform thruster calibration to get the thrust generated by the thruster as a function of the PWM or the voltage signal given to the thruster. However, we circumvented this task by relying on the PID gains only instead of using a separate function to give voltage signals. 
\subsection{Navigation}
Performing localization underwater is a tedious task due to attenuation of GPS signals. In Anahita, the localization of the vehicle is achieved using the Doppler Velocity Log (DVL), Teledyne Pathfinder, and the inertial measurement unit (IMU), Sparton AHRS-8. During tasks, the set points for the trajectory of the vehicle are retrieved from the vision and acoustic systems of the vehicle.\\Once the target is acquired, the coordinates of the vehicle with respect to the target are received and the system tries to achieve goals in that target frame. Task handler layer launches a certain task after the signal from master layer. It receives the values of roll, pitch and yaw from IMU and heave, sway and surge coordinates from hydrophones/vision layer. As all the six PID loops run simultaneously, the vehicle follows the shortest path towards its goal.
\subsection{Mission planner}
The master layer is the topmost layer of our vehicle's software stack. It contains user-fed plan to execute the required tasks by controlling over the bottom layers through service-client calls . For executing a task, it will switch on the vision layer to detect the target and if target is found it will switch on the that task in the task handler layer. The tasks layer also consists of task of separately having surge, sway, heave or yaw motions. \\In between any two tasks combinations of these motions take the vehicle from location of the task it has just ended to a location where it can find the next task. The combination also is set in the master layer by the user. There are switches for all these motions, the main tasks of the competition and vision layer in the master. The switch system makes it very easy to control and make changes in the mission planner.


%%%%%%%%%%%%%%%%%%%%%%%%%%%%%%%%%%%%%%%%%%%%%%%%%%%%%%%%%%%%%%%%%%%%%%%%%%%%%%%%

%%%%%%%%%%%%%%%%%%%%%%%%%%%%%%%%%%%%%%%%%%%%%%%%%%%%%%%%%%%%%%%%%%%%%%%%%%%%%%%%

%%%%%%%%%%%%%%%%%%%%%%%%%%%%%%%%%%%%%%%%%%%%%%%%%%%%%%%%%%%%%%%%%%%%%%%%%%%%%%%%

\subsection{Computer Vision}

\subsubsection {Preprocessing (Blue filter)}

There is poor visibility in underwater vision because of attenuation of the propagated light. The attenuation of light increases exponentially with increasing distance and increasing depth mainly due to absorption and scattering effects. Due to absorption, a raw underwater image appears hazy and the contrast is affected due to scattering. These phenomena are also the reason why distant objects almost disappear, or at the least are difficult to discern, in an underwater image.

We applied these filters in a linear manner:
\includegraphics[width=\linewidth]{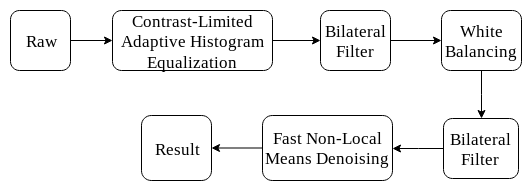}

And these were the results:

\begin{figure}[H]
\centering
\begin{subfigure}[b]{0.4\linewidth}
  \includegraphics[width=\linewidth]{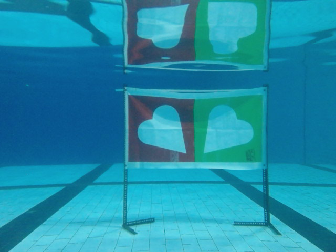}
  \caption{Raw image}
\end{subfigure}
\begin{subfigure}[b]{0.4\linewidth}
  \includegraphics[width=\linewidth]{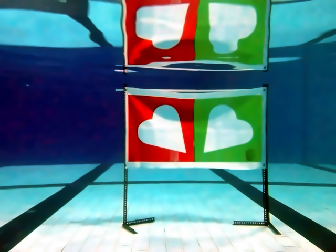}
  \caption{Blue filtered image}
\end{subfigure}
\caption{Pre-processing results}
\end{figure}

\subsubsection*{Contrast-Limited Adaptive Histogram Equalization}

Histogram equalization is a process which equalizes the tonal distribution of the photograph. Adaptive histogram equalization, which works on patches of the image is used because it improves local contrast, rather than global histogram equalization because it heavily changes the brightness of object of interest. The subsequent problem of over-amplification of noise is resolved by contrast-limiting, i.e if any histogram bin is above the specified contrast limit, those pixels are clipped and distributed uniformly to other bins before applying histogram equalization.

\subsubsection*{White Balance}

This operation discards pixel colors at both ends of the histograms generated by the Red, Green and Blue channels of the image to ensure they do not affect the maximal stretching. Which colors are being discarded is decided by the $discard\ ratio$ which is the ratio of the number of pixels showing that color to the total number of pixels. At the end of this process the histogram of each of the three channels covers its entire domain, which tends to reduce the cast the image may earlier have.

\subsubsection{Object Detection}

\begin{figure}[h]
\includegraphics[width=\linewidth]{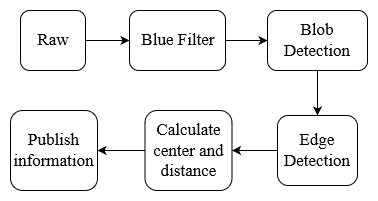}
\caption{Object detection flowchart}
\end{figure}

For most tasks, the raw image is first pre-processed ("blue filtered"). After this, the main object of interest is identified by using techniques like thresholding followed by morphological operations. The edge of this blob is identified either by using Canny Edge detection algorithm followed by Probabilistic Hough Line Transform or contouring.
The center of the blob is calculated using a bounding rectangle, bounding ellipse or by assigning weights to the contours. The distance of the object from the vehicle is calculated with respect to the camera and this is done by an exponential mapping of a particular dimension of the blob with its actual distance from the vehicle.

\begin{figure}[H]
\centering
\begin{subfigure}[b]{0.4\linewidth}
  \includegraphics[width=\linewidth]{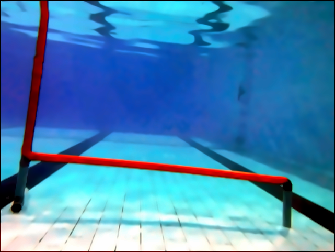}
  \caption{Preprocessed image}
\end{subfigure}
\begin{subfigure}[b]{0.4\linewidth}
  \includegraphics[width=\linewidth]{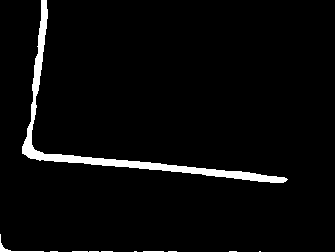}
  \caption{Thresholded and closed}
\end{subfigure}
\begin{subfigure}[b]{0.4\linewidth}
  \includegraphics[width=\linewidth]{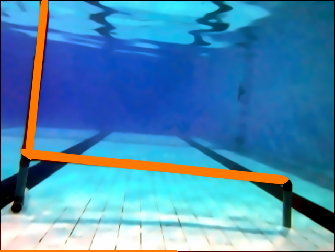}
  \caption{Hough Lines result}
\end{subfigure}
\begin{subfigure}[b]{0.4\linewidth}
  \includegraphics[width=\linewidth]{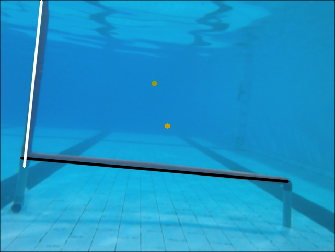}
  \caption{Centre detected}
\end{subfigure}
\caption{Object Detection results}
\end{figure}

% \section*{PRELIMINARY RESULTS}
% Before submitting this paper, the team has completed the design of Anahita, manufactured most of the parts and the whole vehicle will be completed soon. The electrical subsystem has completed the PCBs’ design and will be ready to install in the new vehicle. The new Sparton AHRS-8 and T200 thrusters with ESC modules have been tested and will also be installed in Anahita. The software stack is being tested on the older vehicle, Varun. Every day in the summers testing was done in the institute’s swimming pool for three hours. Buoy, Line, Gate, and Torpedo tasks have been successfully tested and completed by both vision and motion systems.

\section*{ACKNOWLEDGMENT}

We would like to thank DoRD, IIT Kanpur for funding our project. This project would not have been possible without the support of the staff at 4i Lab, Central Workshop and Tinkering Lab at our institute in manufacturing several of our components.\\ The team would also like to extend our gratitude towards the Institute administration and numerous staff members at the swimming pool who supported us during our extended phase of development and testing. We would also like to thank the following sponsors for making the design and fabrication of the AUV possible: ANSYS, Solidworks, Sparton and Mathworks.

\end{document}